\documentclass{svproc}

\usepackage{url}
\usepackage{graphicx}
\usepackage{amsmath}
\usepackage{amssymb}
\usepackage{orcidlink}
\usepackage{hyperref}

\begin{document}
\mainmatter              
\title{A Genetic Fuzzy-Enabled Framework on Robotic Manipulation for In-Space Servicing}
%
\titlerunning{GF-LQR for In-Space Servicing}  
%
\author{Nathan Steffen\inst{1}\orcidlink{0009-0000-1594-3800}
\and Wilhelm Louw\inst{1}\orcidlink{0009-0009-8683-1932}
\and Nicholas Ernest\inst{2}\orcidlink{0000-0002-5819-3541}
\and Timothy Arnett \inst{2}\orcidlink{0000-0001-6759-4593}
\and Kelly Cohen \inst{1}\orcidlink{0000-0002-8655-1465}
}
\authorrunning{Steffen, Louw, Ernest, Arnett, and Cohen} 
%
\tocauthor{Nathan Steffen, Wilhelm Louw, Nick Ernest, Timothy Arnett, Kelly Cohen}
\institute{University of Cincinnati, Cincinnati OH 45221, USA\\
\email{steffenr@mail.uc.edu}, \email{louwwa@ucmail.uc.edu}, \email{cohenky@ucmail.uc.edu}\\
\and Thales Avionics, Blue Ash OH 45242, USA
\email{nick.ernest@defense.us.thalesgroup.com}, \email{tim.arnett@defense.us.thalesgroup.com}}

\maketitle              

\begin{abstract}
Automation of robotic systems for servicing in cislunar space is becoming extremely
important as the number of satellites in orbit increases.
Safety is critical in performing satellite maintenance, so the
control techniques utilized must be trusted, in addition to being highly efficient.
In this work, Genetic Fuzzy Trees are combined with the widely-used LQR control
scheme via Thales' TrUE AI Toolkit\textsuperscript{TM} to create a trusted and
efficient controller for a two-degree-of-freedom planar robotic manipulator
that would theoretically be used to perform satellite maintenance.
It was found that Genetic Fuzzy-LQR is 18.5\% more performant than optimal LQR
on average, and that it is incredibly robust to uncertainty.

\keywords{Space Robotics, Fuzzy Logic, LQR}
\end{abstract}
\section{Introduction}
According to the United Nations Office of Outer Space, there are 11,517 active
registered satellites in Earth's orbit as of 2023, with 8,053 of them being
from the United States alone \cite{unoosa_satellites}. Historically, if a satellite
required servicing, a human astronaut would have to perform that maintenance, or
the satellite would be decommissioned. For the current number of satellites in orbit,
utilizing human operators is not only extremely risky, but also incredibly
impractical. As a result, the use of autonomous robotic systems
that can perform satellite maintenance, refueling, and upgrading is a highly active
area of research \cite{Belvin_2016}\cite{Flores-Abad_Ma_Pham_Ulrich_2014}\cite{Zhu_Ai_Chen_2022}.

Current approaches to automate the procedures required for in-space servicing
are based on artificial intelligence (AI) methods such as deep reinforcement learning (DRL) \cite{Hovell_Ulrich_2021}\cite{Davalos-Guzman_Castañeda_Aguilar-Lobo_Ochoa-Ruiz_2021}\cite{Zappulla_Park_Virgili-Llop_Romano_2019}.
This is due to DRL's ability to perform well in high-dimensional, non-linear environments.
While performance is important for autonomous robotic systems, it is also important to
trust the control methods in place. DRL struggles with trustworthiness as it is difficult
to interpret its output and it becomes unpredictable in situations outside of its
training set. Although there are techniques like run time assurance 
\cite{Hobbs_Mote_Abate_Coogan_Feron_2023}
that help handle unpredictable behavior in controllers, the ability to interpret the system's
actions is still important for interfacing and trusting the system. For this reason, the use of genetic fuzzy systems (GFS) offers value to autonomous space robotics due to
the inherent explainability in the inference process.

In this work, the performance and robustness of a GFS applied to a two-degree-of-freedom
(2DOF) planar robotic manipulator are investigated. This is an
attempt to mimic the robotic arm that would be performing satellite maintenance, and 
serves as a foundation to investigating this approach to other systems that would be 
utilized for in-orbit servicing. Here, the specific implementation utilizes the 
optimality guarantee of the Linear Quadratic Regulator (LQR) feedback control scheme 
where the GFS generates the weights used in solving the continuous algebraic Ricatti 
equation. Once trained, the performance of this controller is then compared to optimal
static LQR gains which were optimized using the same fitness function as the hybrid
Genetic Fuzzy-LQR controller. This approach builds off other works, such as
\cite{Hernandez-Pineda_Bezerra-Viana_Marques-Simoes_Carvalho_Bezerra_2023},
where a hybrid Genetic Fuzzy-LQR approach is applied to a smaller robotic manipulator and the rule 
base is not optimized, and \cite{Al-Mola_Abdelmaksoud_2023}, where a GFS is applied to a 
PID controller instead.

\section{Manipulator Dynamics}
\begin{figure}[htb]
	\centering\includegraphics[width=3.0in]{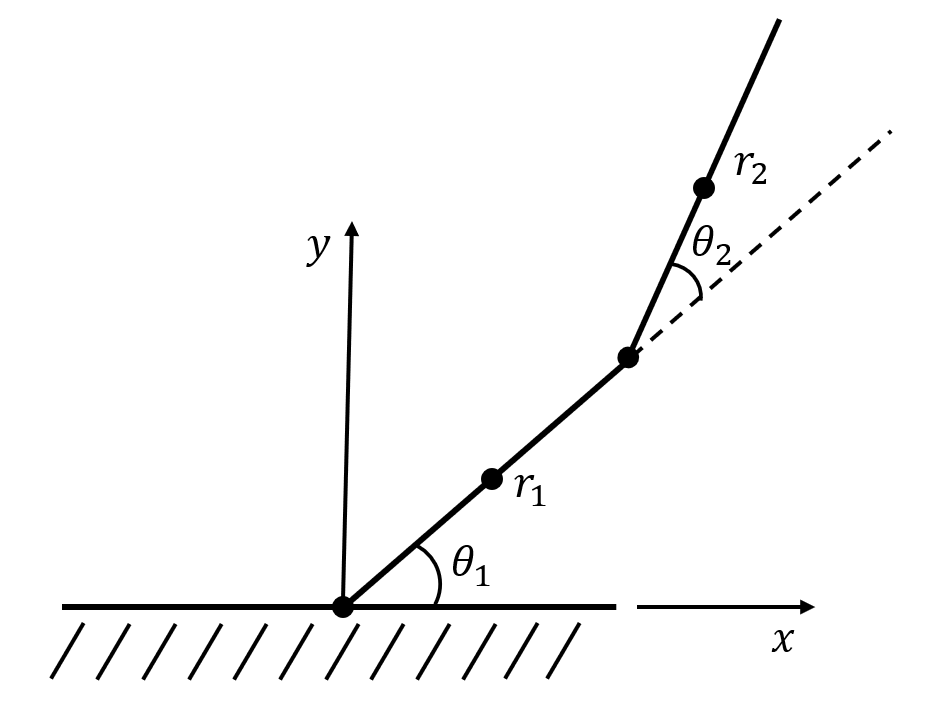}
	\caption{Representation of 2DOF Arm System.}
	\label{fig:1}
\end{figure}

A representation of the system being examined can be seen in Figure \ref{fig:1}. Here,
$r_1$ and $r_2$ are the distances along the links that are considered as the positions, 
and $\theta_1$ and $\theta_2$ are the angles of the links. The position expressions can
then be seen below.

\begin{equation}
    \mathbf{x_1} = \left[r_1\cos{\theta_1}, \ r_1\sin{\theta_1}\right]^T
    \label{pos:1}
\end{equation}

\begin{equation}
\mathbf{x_2} = \left[l_1\cos{\theta_1} + r_2\cos{\left(\theta_1+\theta_2\right)},  \ l_1\sin{\theta_1} + r_2\sin{\left(\theta_1+\theta_2\right)}\right]^T
\label{pos:2}
\end{equation}
\\
In Equation \ref{pos:2}, $l_1$ is the total length of link 1. Having expressions for the 
positions of the links, the velocities can be found by taking the time derivative of
these equations.

\begin{equation}
    \mathbf{v_i} = \frac{d\mathbf{x_i}}{dt} \ for \ i=1,2 
    \label{vel}
\end{equation}

The generalized coordinates for the system are the angles of the links, $\theta_1$ and 
$\theta_2$. With this in mind, the Lagrangian $L=T-V$ can be formed by finding the kinetic 
and potential energy of the system.

\begin{equation}
    T = \frac{1}{2}m_1\mathbf{v_1}^T\mathbf{v_1} + \frac{1}{2}m_2\mathbf{v_2}^T\mathbf{v_2} + \frac{1}{2}I_1\omega_1^2 + \frac{1}{2}I_2\omega_2^2
    \label{ke}
\end{equation}

Here, $m_1$ and $m_2$ are the masses of the links, $I_1$ and $I_2$ are the moments of 
inertia of the links, and $\omega_1$ and $\omega_2$ are $\dot{\theta}_1$ and 
$\dot{\theta}_2$ respectively. The potential energy of the system, $V$, is simply 0 since
the robotic system is in space, and thus gravitational effects are considered negligible.

With the Lagrangian found, the equations of motion can be solved for. These take the form below.

\begin{equation}
    \mathbf{M}(\mathbf{\theta})\mathbf{\ddot{\theta}} + \mathbf{C}(\mathbf{\theta}, \mathbf{\dot{\theta}}) + \mathbf{G}(\mathbf{\theta}) = \mathbf{\tau}
\end{equation}

Here, $\mathbf{M}(\mathbf{\theta}) \in \mathcal{R}^{2\times2}$ is the generalized inertial matrix,
$\mathbf{C}(\mathbf{\theta}, \mathbf{\dot{\theta}}) \in \mathcal{R}^{2\times1}$ is the Coriolis vector, $\mathbf{G}(\mathbf{\theta}) \in \mathcal{R}^{2\times1}$
is the gravitational force vector, and $\mathbf{\tau} \in \mathcal{R}^{2\times1}$ is the control vector. The 
generalized inertial matrix element-wise in the equations below.

\begin{equation}
    M_{11} = m_1r_1^2 + m_2\left(l_1^2+r_2^2\right) + I_1 + m_2l_1r_2\cos{\theta_2}
\end{equation}

\begin{equation}
    M_{12} = M_{21} = m_2r_2^2 + \frac{1}{2}m_2l_1r_2\cos{\theta_2}
\end{equation}

\begin{equation}
    M_{22} = m_2r_2^2 + I_2
\end{equation}

The elements of the Coriolis vector are then defined as

\begin{equation}
    C_1 = -\frac{1}{2}m_2l_1r_2\dot{\theta_2}\left(2\dot{\theta}_1+1\right)\sin{\theta_2}
\end{equation}

\begin{equation}
    C_2 = -\frac{1}{2}m_2l_2r_2\dot{\theta_1}\dot{\theta_2}\sin{\theta_2}
\end{equation}

Finally, the gravitational force vector is considered to be negligible here since this application
occurs in space. 

\section{Controller Design}
\subsection{Linear Quadratic Regulator}
The Linear Quadratic Regulator (LQR) is an optimal control scheme that is typically used in linear 
dynamic systems. For non-linear systems, a linearized form of the dynamics is used as an approximation
for the LQR controller. LQR seeks to minimize the infinite-horizon quadratic cost function seen below.

\begin{equation}
    J = \int_0^\infty \left[\mathbf{x}^T\mathbf{Qx} + \mathbf{u}^T\mathbf{Ru}\right] \ dt
    \label{lqr_cost}
\end{equation}

In Equation \ref{lqr_cost}, $\mathbf{x}$ is the state vector, $\mathbf{u}$ is the control vector,
$\mathbf{Q}$ is the weighting matrix that penalizes state deviations, and $\mathbf{R}$ is the weighting 
matrix that penalizes the magnitude of the control inputs. $\mathbf{Q}$ and $\mathbf{R}$ are typically
diagonal matrices, and the values along the diagonals are chosen such that the controller designer
can prioritize state deviations over control effort, or vice versa. 

The optimal control is then given by considering the state-feedback law:

\begin{equation}
    \mathbf{K} = \mathbf{R}^{-1}\mathbf{B}^T\mathbf{P}
    \label{eq:K}
\end{equation}

Here, $\mathbf{P}$ is the solution to the continuous-time algebraic Ricatti equation.

\begin{equation}
    \mathbf{A}^T\mathbf{P} + \mathbf{PA} - \mathbf{PBR}^{-1}\mathbf{B}^T\mathbf{P} + \mathbf{Q} = 0
    \label{eq:CARE}
\end{equation}

Solving Equation \ref{eq:CARE} yields all the necessary information for Equation \ref{eq:K}, and so 
a stable closed-loop control system can be applied\cite{Ogata_2010}.

\subsection{Hybrid Genetic Fuzzy-LQR Controller}
While LQR makes it simple for a controller designer to specify an optimal control for a given problem,
it is still a linear controller so its behavior in a non-linear environment may be suboptimal. To utilize
LQR's capability in a non-linear environment, Fuzzy Logic \cite{Zadeh_1964} can be used to learn the appropriate $\mathbf{Q}$
and $\mathbf{R}$ gains given the states of the system. Essentially, the Fuzzy Inference System (FIS) can
learn when it is best to prioritize achieving a state, and when it is best to prioritize saving control
effort.

In this work, the $\mathbf{R}$ gains are kept constant to lower the solution space, where $r=0.0001$ and 
$\mathbf{R}=\text{diag}(r, r)$. It is assumed that varying the priority in states by 
generating $\mathbf{Q}$ gains will introduce control preserving benefits.

\begin{figure}[htb]
	\centering\includegraphics[width=3.0in]{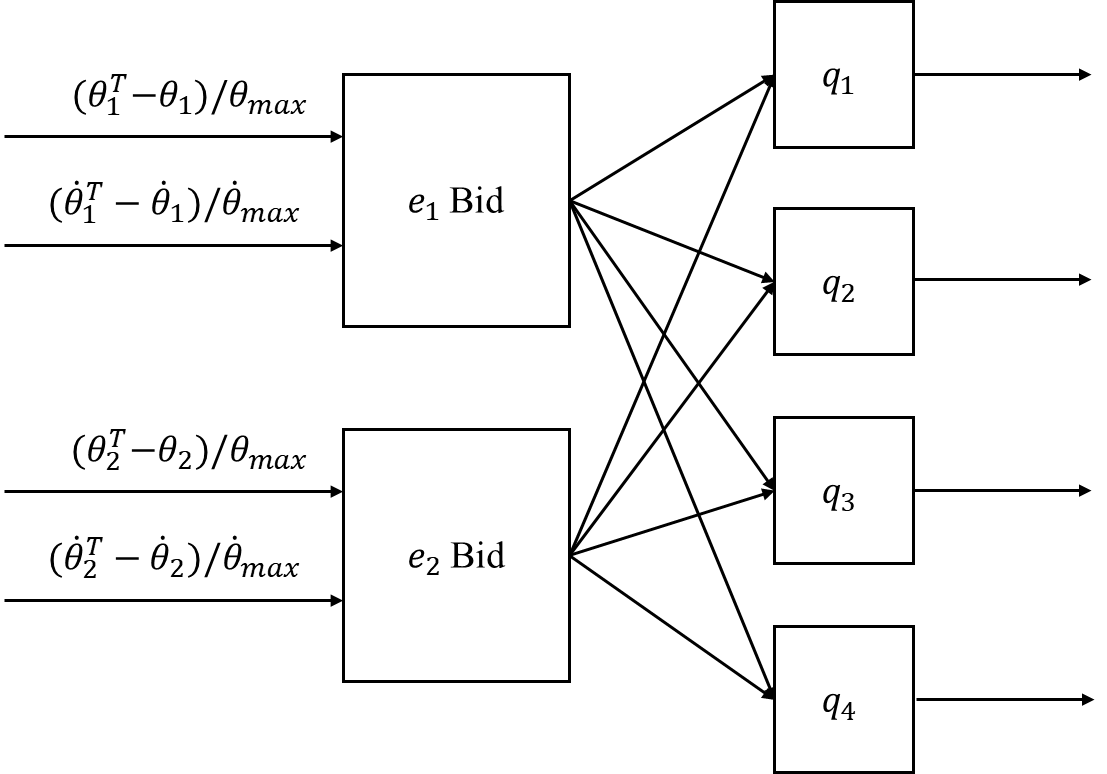}
	\caption{Genetic Fuzzy Tree for Interfacing with LQR.}
	\label{fig:2}
\end{figure}

The method for generating the $\mathbf{Q}$ gains utilizes a Genetic Fuzzy Tree (GFT), as 
seen in Figure \ref{fig:2}. The first layer is intended to create a prioritization ranking
for the two joints, refered to as a ``bid''. As an input to these FISs,
the normalized error and normalized time derivative error inform the bid,
which is a value on the range $[-1, 1]$. This value is then passed to each $\mathbf{Q}$
gain FIS, and each individual FIS rulebase is used to infer the proper gain at that timestep.
$\mathbf{Q}$ is then defined as $\mathbf{Q}=\text{diag}(q_1,q_2,q_3,q_4)$, and Equations
\ref{eq:K} and \ref{eq:CARE} can be used to calculate the feedback gains to produce the
proper control.

To implement the GFT in this way, Thales' TrUE AI Toolkit\textsuperscript{TM} 
\cite{Thales_Group_2019} was used for its FIS framework. TrUE AI\textsuperscript{TM} is a
commercial SDK utilizing Genetic Fuzzy Systems for AI system development. It allowed
the GFT in Figure \ref{fig:2} to be implemented with ease, and offered a streamlined
means to interface for optimization.
Each FIS is defined for Takagi-Sugeno-Kang defuzzification 
\cite{Zhang_Wang_Zhou_Huang_Lam_Sheng_Choi_Cai_Ding_2024}.

\subsection{Optimization Approach}
As implied with the ``Genetic'' part of GFT, the Genetic Algorithm (GA) was used to optimize 
the controller. This is an evolutionary-inspired algorithm that takes principles from the 
theory of natural selection to converge upon a solution that optimizes some fitness function.
The solution comes in the form of a chromosome, which is typically an array of integers that
a user is capable of encoding and decoding\cite{Sastry2005}.

The specific GA implementation used was also part of Thales' TrUE AI Toolkit\textsuperscript{TM}. 
While typically there are a number of hyper-parameters that could be tuned for a GA
(e.g. mutation rate, elitism, etc.), TrUE AI\textsuperscript{TM} is capable of
intelligently determining the proper hyper-parameters on its own. All that must be specified
are the population size, the number of populations, the number of generations, and the
fitness function.

For training a hybrid Genetic Fuzzy-LQR controller, the primary focus was to minimize both
settling time and control effort. This takes form in the fitness function in Equation
\ref{eq:cost}, where $T_s$ is the settling time and $IAC_i$ is the integral over
the absolute control for joint $i$, defined in Equation \ref{eq:IAC}. In this case, the
``fitness'' function is really a cost function, since a minimum cost is desired rather than 
a minimum fitness.

\begin{equation}
    Cost=T_s+\frac{IAC_1}{\tau_{1_{max}}}+\frac{IAC_2}{\tau_{2_{max}}}
    \label{eq:cost}
\end{equation}

\begin{equation}
    IAC_i=\sum_{j=1}^{N} |\tau_i(t_j)|\times(t_j-t_{j-1}) \ \ \text{for} \ i=1,2
    \label{eq:IAC}
\end{equation}

Here, $\tau_{i_{max}}$ is a user-defined value representative of the maximum torque a
joint can produce for the system, and $N$ is the number of time steps for the propagation
to achieve its settling time.

It is important to keep in mind that this is the cost for a controller over a single set of
initial conditions and target states. So, if multiple cases are observed, the overall cost is
shown in Equation \ref{eq:overall_cost}, where $M$ is the number of cases being examined.

\begin{equation}
    Overall \ Cost=\sum_{i=0}^{M}\left(T_{s_i} + \frac{IAC_{1_i}}{\tau_{1_{max}}}+\frac{IAC_{2_i}}{\tau_{2_{max}}} \right)
    \label{eq:overall_cost}
\end{equation}

\section{Results \& Discussion}
\subsection{Training Setup}

\setlength{\textfloatsep}{10pt} 
\begin{table}[h!]
\centering
\caption{Manipulator Parameters.}
\label{tb:1}
\begin{tabular}{r@{\quad}rl}
\hline
\multicolumn{1}{l}{Parameter} & \multicolumn{2}{l}{Value} \\
\hline
$m_1$  & 20 kg & \\
$m_2$  & 10 kg & \\
$l_1$  & 1 m   & \\
$l_2$  & 1 m   & \\
\hline
\end{tabular}
\end{table}

Table \ref{tb:1} shows the values used to define the 2DOF planar robotic manipulator being
investigated. $r_i$ for $i=1,2$ is considered to be half the length of its respective arm,
and the moments of inertia are defined as $I_i=\frac{1}{3}m_il_i^2$. For this system, 
$\tau_{1_{max}}$ was defined as 400 Nm, and $\tau_{2_{max}}$ was defined as 150 Nm. This
was intended to keep $\ddot{\theta}_i\le 5 \ \text{degrees/second}^2$, or at least in that range.

The training of the Genetic Fuzzy-LQR utilized 88 sets of initial conditions and target states.
44 of these scenarios used an initial condition with $\theta_1=0^o$ and $\theta_2=0^o$, and the
other 44 used an initial condition with $\theta_1=180^o$ and $\theta_2=0^o$. Both sets of initial
conditions had the arm begin at rest. Each target state was an increment of $45^o$ in both joints.
The first joint was limited to the range $[0^o, 180^o]$ for these increments, whereas the second
joint utilized the range $[-180^o, 180^o]$.

Each simulation utilized Runge-Kutta 4 numeric integration at a constant timestep of 0.0167 seconds.
The maximum allowable time that propagation can occur is 10 seconds, and the propagation is ended
if the difference between every state is less than or equal to some tolerance. Here, the tolerance
is set to be 0.02.

Prior to training the Genetic Fuzzy-LQR, optimal LQR feedback gains were found for each of these
cases using the cost function in Equation \ref{eq:cost}. Once determined, the optimal LQR costs
were cached, and then were utilized in a new cost function per case for the Genetic Fuzzy-LQR, seen in 
Equation \ref{eq:new_cost}.

\begin{equation}
    Cost=\left(T_s+\frac{IAC_1}{\tau_{1_{max}}}+\frac{IAC_2}{\tau_{2_{max}}}\right) / optimal \ LQR \ cost
    \label{eq:new_cost}
\end{equation}

It is important to keep in mind that this is the cost per case. Similar to Equation \ref{eq:overall_cost},
if a controller's cost over multiple cases is to be found, simply sum those costs.

Now, the Genetic Fuzzy-LQR controller can be trained directly against the optimal LQR gains for
every scenario. For the overall training metric, the average cost over cases is utilized. So,
once each case's cost is summed together, that value is simply divided by 88 to get the overall
cost of the controller.

As was mentioned before, TrUE AI\textsuperscript{TM} only requires population size, number of
populations, and number of generations for its GA. For this training, 112 chromosomes were used with
4 populations for 1,500 generations. The first layer in the GFT (the error bids) was selected to 
have 3 membership functions per FIS, and the second layer ($\mathbf{Q}$ gain evaluation) was
selected to have 7 membership functions. The number of rules in each FIS is the number of 
membership functions squared. For the following training, only the rules and output bounds were
optimized, while the membership function shapes were kept static.

\subsection{Training Results}
\begin{figure}[htb]
	\centering\includegraphics[width=3.0in]{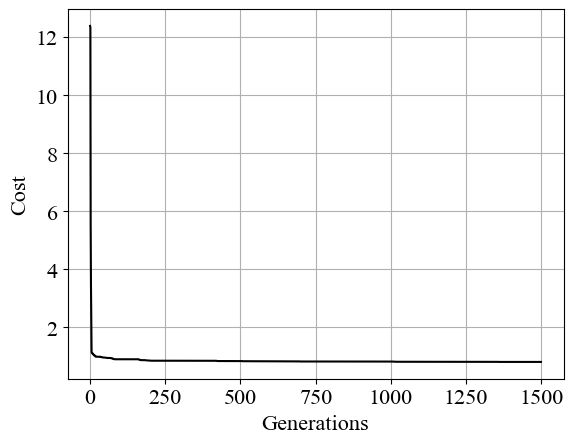}
	\caption{Training History.}
	\label{fig:training}
\end{figure}

Figure \ref{fig:training} shows the Genetic Fuzzy-LQR's cost minimization history. The value on which it 
ended was 0.815. Since this is the average cost over each case, this means that the Genetic Fuzzy-LQR
controller is around 18.5\% more performant on average than optimal LQR in each case. While this may make it
seem like Genetic Fuzzy-LQR is better in every case, comparing the costs of each case directly shows
that Genetic Fuzzy-LQR outperforms optimal LQR in only 86 out of 88 cases. These two cases showed
marginally better costs for optimal LQR than for Genetic Fuzzy-LQR, and so it may be possible to 
overcome this difference with more training.

\begin{figure}[htbp]
\centering
\begin{tabular}{ccc}
\includegraphics[width=0.3\textwidth]{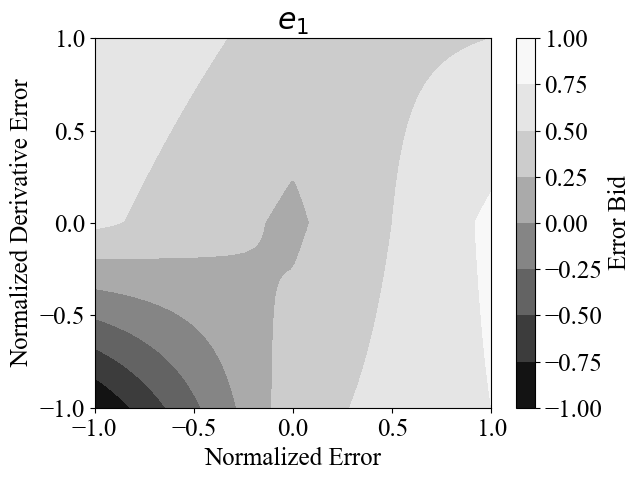} &
\includegraphics[width=0.3\textwidth]{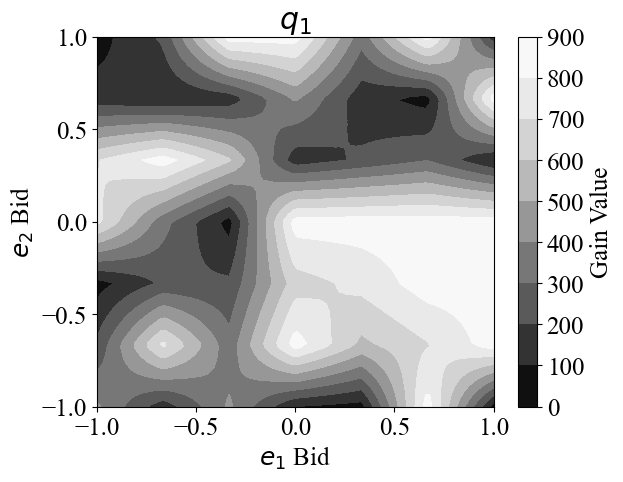} &
\includegraphics[width=0.3\textwidth]{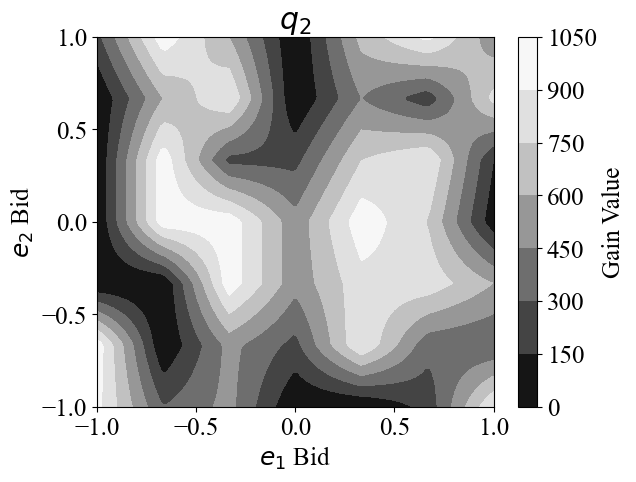} \\
\includegraphics[width=0.3\textwidth]{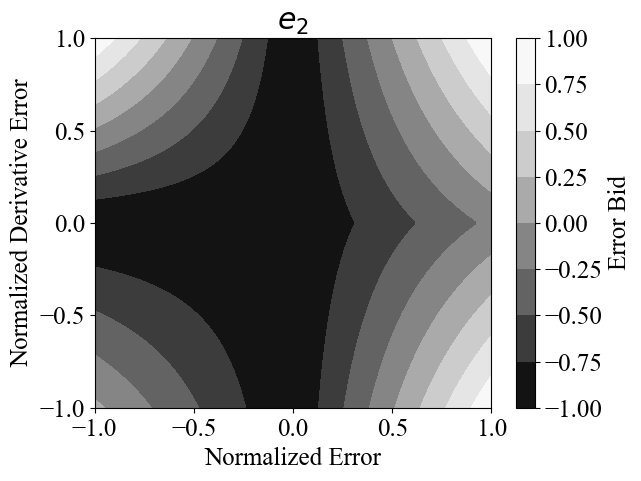} &
\includegraphics[width=0.3\textwidth]{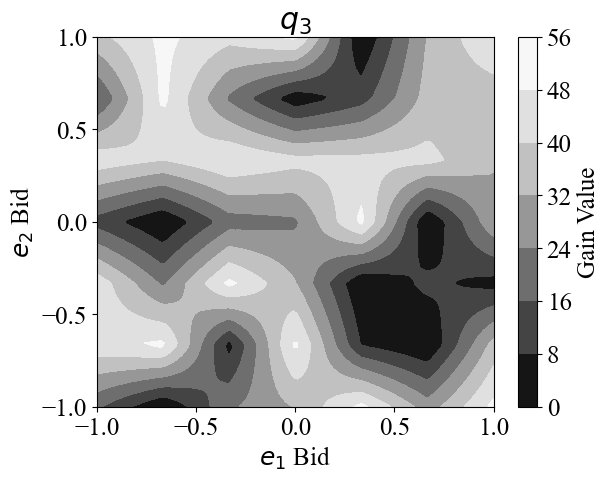} &
\includegraphics[width=0.3\textwidth]{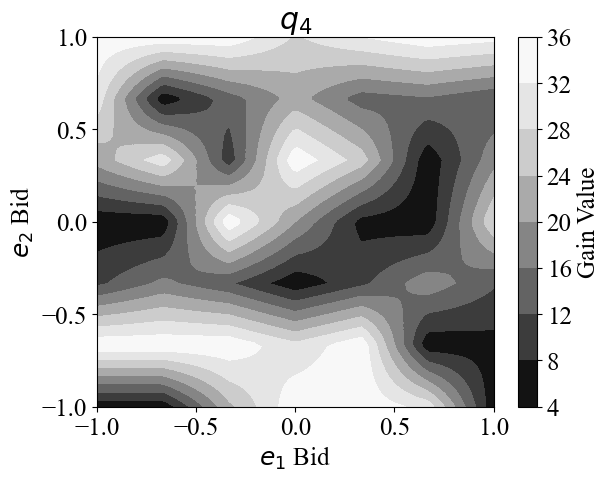} \\
\end{tabular}
\caption{FIS Control Surfaces.}
\label{fig:surfaces}
\end{figure}

The resulting FIS control surfaces can be seen in Figure \ref{fig:surfaces}.
Since the first layer utilizes 3 membership functions per FIS, it can be seen how much simpler its
surface is compared to the 7 membership functions per FIS used in the second layer. Additionally,
since the output bounds were trained with the rules, it is worth noting that the color scales for
the $\mathbf{Q}$ gain FISs are different. Here, the $q_1$ and $q_2$ bounds cover a wider range as
the GA learned that the $\theta_1$ and $\theta_2$ gains affect the system more than the 
$\dot{\theta}_1$ and $\dot{\theta}_2$ gains do.

\subsection{Single Case Comparison}
\begin{figure}[htbp]
\centering
\begin{tabular}{ccc}
\includegraphics[width=0.3\textwidth]{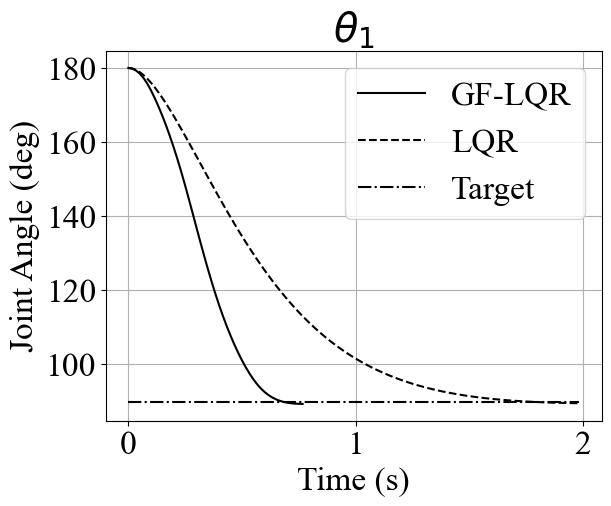} &
\includegraphics[width=0.3\textwidth]{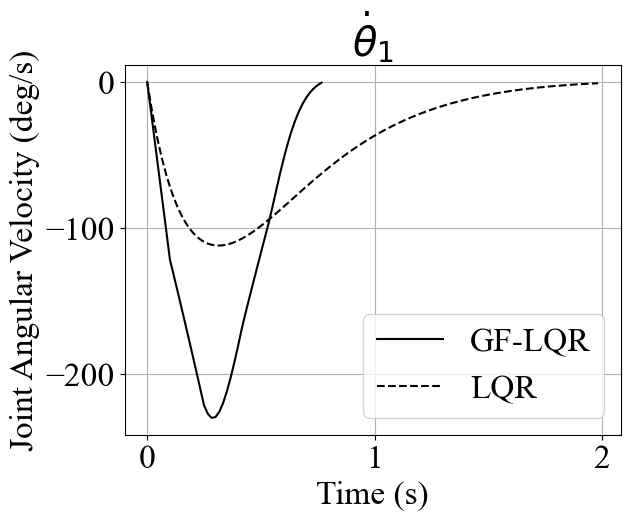} &
\includegraphics[width=0.3\textwidth]{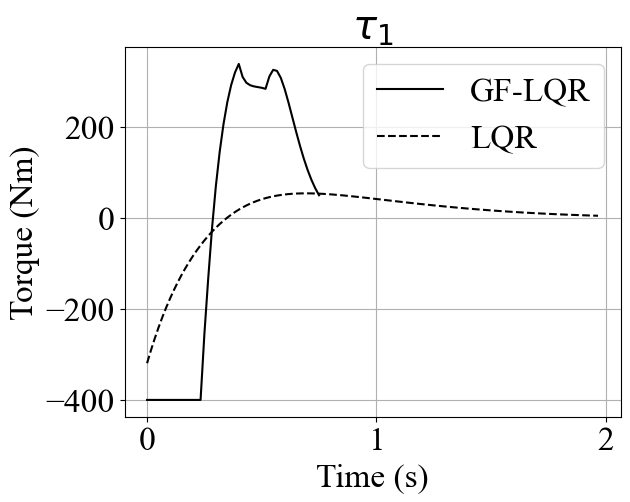} \\
\includegraphics[width=0.3\textwidth]{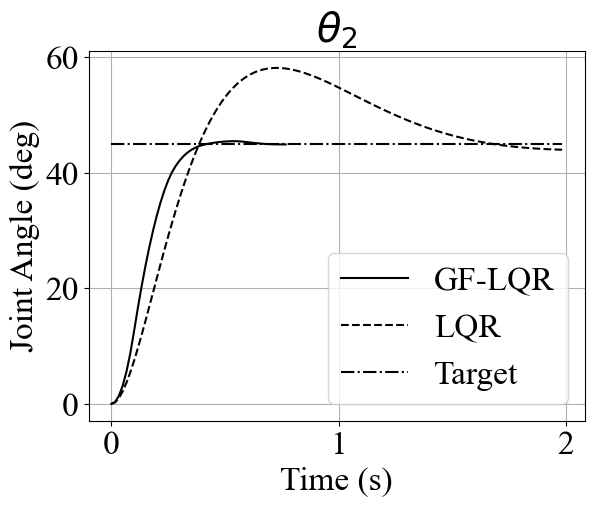} &
\includegraphics[width=0.3\textwidth]{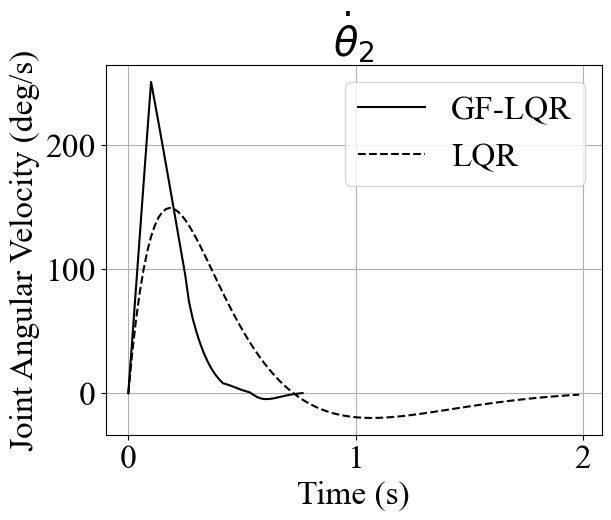} &
\includegraphics[width=0.3\textwidth]{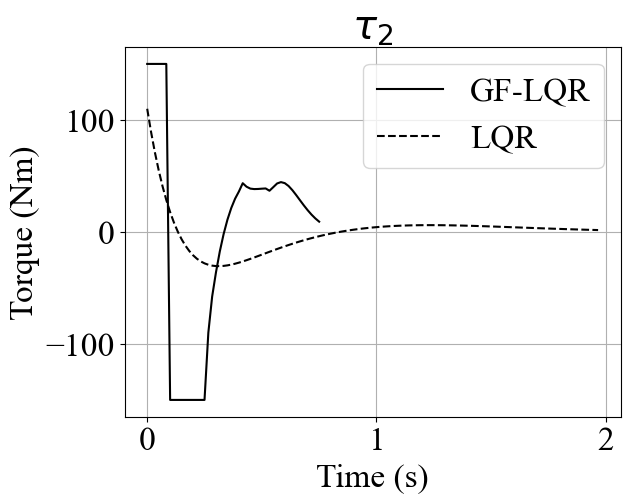} \\
\end{tabular}
\caption{Single Case State and Control Comparison.}
\label{fig:comparison}
\end{figure}

Here, a single case is compared between optimal LQR and Genetic Fuzzy-LQR to see the difference 
in control behavior. In the case shown in Figure \ref{fig:comparison}, the initial state is
$\theta_1=180^o$ and $\theta_2=0^o$, and the target state is $\theta_1=90^o$ and $\theta_2=45^o$.

Clear non-linear behavior can be seen in the control plot for this example, especially in 
comparison to the optimal LQR's controller behavior. Overall, this results in Genetic Fuzzy-LQR
having a quicker settling time, but perhaps at the cost of too much control effort. Regardless,
this controller outperforms optimal LQR in terms of the defined cost function, so the control
behavior can be tuned to more desirable behavior if needed.

\begin{figure}[htbp]
\centering
\begin{tabular}{ccc}
\includegraphics[width=0.45\textwidth]{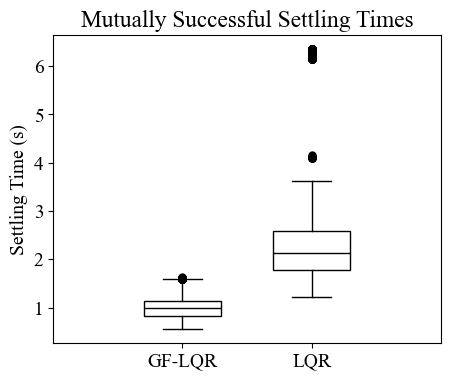} &
\includegraphics[width=0.45\textwidth]{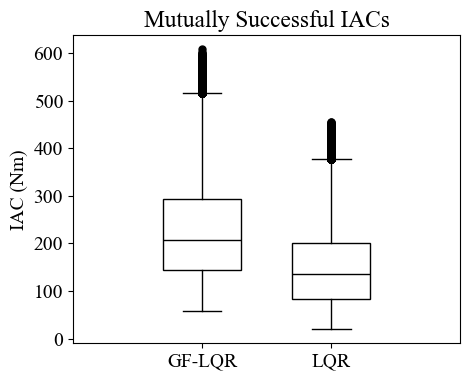} & \\
\multicolumn{3}{c}{\includegraphics[width=0.45\textwidth]{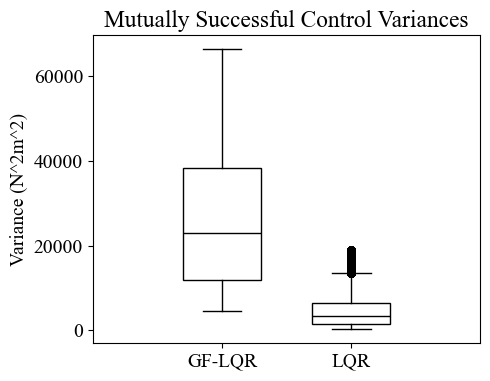}} \\\end{tabular}
\caption{Robustness Analysis Statistics.}
\label{fig:ra_stats}
\end{figure}

\subsection{Robustness Analysis}
The robustness of a controller to uncertainty in space robotic systems is of utmost importance.
While in orbit, a satellite is susceptible to extreme temperatures and measurement uncertainties,
among other risks, but despite this, the controller still needs to perform well. To test the
Genetic Fuzzy-LQR's robustness to uncertainty compared to optimal LQR, 1,001,088 simulations were
ran where mass and length varied uniformly randomly by +/- 10\%.

Out of these cases, Genetic Fuzzy-LQR was successful in reaching the target state 100\% of the time,
while optimal LQR only reached its target state 89.8\% of the time. Figure \ref{fig:ra_stats} shows the
settling time, IAC, and control variance statistics of the mutually successful cases to compare
controller reliability.

It can be seen in the top left graph of Figure \ref{fig:ra_stats} that Genetic Fuzzy-LQR is capable
of achieving its target state much faster than optimal LQR under uncertainty. When considering its
100\% success rate, this indicates it is not only highly performant, but also extremely reliable.

When considering Genetic Fuzzy-LQR's control behavior over these simulations, however, it does
appear to utilize vastly more control effort to make up for the uncertainty. This can be seen
in the distributions present in the top right and bottom graphs in Figure \ref{fig:ra_stats},
and again, this behavior can likely be tuned within the cost function for better control effort
preservation.

\section{Conclusions}
In this work, a Genetic Fuzzy Trees were combined with the Linear Quadratic Regulator control
scheme to produce a non-linear controller that performs extremely well and is robust to uncertainty.
It was found that in every case examined, the Genetic Fuzzy-LQR outperformed optimal LQR according
to the previously defined cost function. For future work, more cost functions should be examined
for better control behavior, more degrees of freedom should be added with an appropriate GFT
architecture, and more verification techniques should be explored for this system to increase
its overall trustworthiness.

\section{Acknowledgments}
The authors express their appreciation to Thales for providing licenses to utilize its TRUE AI framework. Without this assistance, the development of this controller would have been significantly more arduous and the outcomes would have been substantially diminished.

Additionally, the authors extend their sincere gratitude to the members of the AI Bio Lab at the University of
Cincinnati for their invaluable discussions and collaborative efforts that facilitated the
realization of this work. In particular, the contributions of Magnus Sieverding, Jared Burton,
Lohith Pentapalli, Bharadwaj Dogga, Tri Nguyen, Garret Olges, Kaus Shankar, Lucia Vilar Nuño,
and Hugo Henry are highly appreciated.

Finally, we acknowledge the support provided by the Ohio Space Grant Consortium, which has 
provided the requisite resources and funding to enable this research.


\begin{thebibliography}{15}
\bibitem{unoosa_satellites}
United Nations Office for Outer Space Affairs:
Satellite Database.
\url{https://www.unoosa.org/oosa/osoindex/}.
Accessed: 28 February 2025 (2023)

\bibitem{Belvin_2016}
W.~K. Belvin, W.~R. Doggett, J.~C. Watson, E.~E. Komendera, T.~Mann, and L.~M. Bowman:
In-space structural assembly: Applications and technology.
In: \emph{3rd AIAA Spacecraft Structures Conference} (Jan.\ 2016)

\bibitem{Flores-Abad_Ma_Pham_Ulrich_2014}
A.~Flores‑Abad, O.~Ma, K.~Pham, and S.~Ulrich:
A review of Space Robotics Technologies for on‑orbit servicing.
\emph{Progress in Aerospace Sciences} \textbf{68} (Jul.\ 2014), 1–26.
DOI:\url{10.1016/j.paerosci.2014.03.002}

\bibitem{Zhu_Ai_Chen_2022}
A.~Zhu, H.~Ai, and L.~Chen:
A fuzzy logic reinforcement learning control with spring‑damper device for space robot capturing satellite.
\emph{Applied Sciences} \textbf{12}(5) (Mar.\ 2022), Art.\ no.\ 2662.
DOI:\url{10.3390/app12052662}

\bibitem{Hovell_Ulrich_2021}
K.~Hovell and S.~Ulrich:
Deep Reinforcement Learning for Spacecraft Proximity Operations Guidance.
\emph{Journal of Spacecraft and Rockets} \textbf{58}(2) (Mar.\ 2021), 254–264.
DOI:\url{10.2514/1.a34838}

\bibitem{Davalos-Guzman_Castañeda_Aguilar-Lobo_Ochoa-Ruiz_2021}
U.~Davalos\-Guzman, C.~E. Castañeda, L.~M. Aguilar\-Lobo, and G.~Ochoa\-Ruiz:
Design and implementation of a real time control system for a 2DOF robot based on recurrent high order neural network using a hardware in the loop architecture.
\emph{Applied Sciences} \textbf{11}(3) (Jan.\ 2021), Art.\ no.\ 1154.
DOI:\url{10.3390/app11031154}

\bibitem{Zappulla_Park_Virgili-Llop_Romano_2019}
R.~Zappulla, H.~Park, J.~Virgili‑Llop, and M.~Romano:
Real‑time autonomous spacecraft proximity maneuvers and docking using an adaptive artificial potential field approach.
\emph{IEEE Transactions on Control Systems Technology} \textbf{27}(6) (Nov.\ 2019), 2598–2605.
DOI:\url{10.1109/tcst.2018.2866963}

\bibitem{Hobbs_Mote_Abate_Coogan_Feron_2023}
K.~L. Hobbs, M.~L. Mote, M.~C.~L. Abate, S.~D. Coogan, and E.~M. Feron:
Runtime assurance for safety-critical systems: An introduction to safety filtering approaches for complex control systems.
\emph{IEEE Control Systems} \textbf{43}(2) (Apr.\ 2023), 28–65.
DOI:\url{10.1109/mcs.2023.3234380}

\bibitem{Hernandez-Pineda_Bezerra-Viana_Marques-Simoes_Carvalho_Bezerra_2023}
A.~Hernandez‑Pineda, I.~Bezerra‑Viana, M.~Marques‑Simoes, and F.~Carvalho~Bezerra:
LQR combined with fuzzy control for 2‑DOF planar robot trajectories.
\emph{Proceedings of the 20th International Conference on Informatics in Control, Automation and Robotics} (2023), 763–769.
DOI:\url{10.5220/0012258600003543}

\bibitem{Al-Mola_Abdelmaksoud_2023}
M.~H. Al‑Mola and S.~I. Abdelmaksoud:
Performance improvement of a space robotic manipulator using Fuzzy‑based PID controller.
\emph{2023 International Telecommunications Conference (ITC‑Egypt)} (Jul.\ 2023), 524–529.
DOI:\url{10.1109/itc-egypt58155.2023.10206133}

\bibitem{Ogata_2010}
K.~Ogata:
\emph{Modern Control Engineering}.
Pearson, Boston, United States (2010).

\bibitem{Zadeh_1964}
L.~A. Zadeh:
Fuzzy Sets.
\emph{Information and Control}, Nov.\ 1964.
DOI:\url{10.21236/ad0608981}

\bibitem{Thales_Group_2019}
Thales Group:
Thales' True AI Approach to Artificial Intelligence to be Unveiled in Paris.
\emph{Thales Group}, Jun. 2019.
\url{https://www.thalesgroup.com/en/group/journalist/press_release/thales-true-ai-approach-artificial-intelligence-be-unveiled-paris}

\bibitem{Zhang_Wang_Zhou_Huang_Lam_Sheng_Choi_Cai_Ding_2024}
Y.~Zhang, G.~Wang, T.~Zhou, J.~Huang, S.~Lam, K.~C. Choi, J.~Cai, and W.~Ding:
Takagi–Sugeno–Kang Fuzzy System Fusion: A survey at hierarchical, wide and stacked levels.
\emph{Information Fusion} \textbf{101} (Jan.\ 2024), Art.\ no.\ 101977.
DOI:\url{10.1016/j.inffus.2023.101977}

\bibitem{Sastry2005}
K.~Sastry, D.~Goldberg, and G.~Kendall:
Genetic Algorithms.
In: \emph{Search Methodologies: Introductory Tutorials in Optimization and Decision Support Techniques}, pp.\ 97–125, Springer US, Boston, MA (2005).
DOI:\url{10.1007/0-387-28356-0_4}


\end{thebibliography}

\end{document}